\documentclass{article}

\usepackage[preprint]{neurips_2021}

\usepackage[utf8]{inputenc} 
\usepackage[T1]{fontenc}    
\usepackage{hyperref}       
\usepackage{url}            
\usepackage{booktabs}       
\usepackage{amsfonts}       
\usepackage{nicefrac}       
\usepackage{microtype}      
\usepackage{xcolor}         
\usepackage{graphicx}
\usepackage{scalefnt}
\usepackage{natbib}

\usepackage{subfig}

\usepackage{adjustbox}
\usepackage{array}

\newcolumntype{R}[2]{%
    >{\adjustbox{angle=#1,lap=\width-(#2)}\bgroup}%
    l%
    <{\egroup}%
}
\newcommand*\rot{\multicolumn{1}{R{45}{1em}}}

\newcommand\blfootnote[1]{%
  \begingroup
  \renewcommand\thefootnote{}\footnote{#1}%
  \addtocounter{footnote}{-1}%
  \endgroup
}

\newcommand{\benchname}{\textit{SocialAI}}

\title{SocialAI: Benchmarking Socio-Cognitive Abilities in Deep Reinforcement Learning Agents}

\author{Grgur Kova\v c${^{*}}{^{\dagger}}$ \\
        Inria (FR) \\
  
  \And
  
  Rémy Portelas${^{*}}{^{\dagger}}$ \\
  Inria (FR) \\

  \And
  
  Katja Hofmann \\
  Microsoft Research (UK) \\
  
  \And
  
  Pierre-Yves Oudeyer \\
  Inria (FR) \\
}

\begin{document}

\maketitle

\begin{abstract}

Building embodied autonomous agents capable of participating in social interactions with humans is one of the main challenges in AI.
Within the Deep Reinforcement Learning (DRL) field, this objective motivated multiple works on embodied language use.
However, current approaches focus on language as a communication tool in very simplified and non-diverse social situations: the "naturalness" of language is reduced to the concept of high vocabulary size and variability. In this paper, we argue that aiming towards human-level AI requires a broader set of key social skills: 1) language use in complex and variable social contexts; 2) beyond language, complex embodied communication in multimodal settings within constantly evolving social worlds.
We explain how concepts from cognitive sciences could help AI to draw a roadmap towards human-like intelligence, with a focus on its social dimensions. As a first step, we propose to expand current research to a broader set of core social skills.
To do this, we present \benchname, a benchmark to assess the acquisition of social skills of DRL agents using multiple grid-world environments featuring other (scripted) social agents. We then study the limits of a recent SOTA DRL approach when tested on \benchname~and discuss important next steps towards proficient social agents.
Videos and code are available at {\small{\url{https://sites.google.com/view/socialai}}}.
\end{abstract}

\section{Introduction}
\label{sec:introduction}
\blfootnote{$^{*}$Equal contribution}
\blfootnote{$^{\dagger}$Email grgur.kovac@inria.fr \& remy.portelas@inria.fr}

How do human children manage to reach the social and cognitive complexity of human adults? For Vygotsky, a soviet scholar from the 1920s, a main driver for this path towards "higher-level" cognition are socio-cultural interactions with other human beings \citep{vygot-book}. For him, many high-level cognitive functions a child develops first appear at the social level and then develop at the individual level. This leap from interpersonal processes to intrapersonal processes is referred to as \textit{internalization}. Vygotsky's theories influenced multiple works within cognitive science \citep{clark-being-there,hutchins96a}, primatology \citep{tomasello-culture-cognition} and the developmental robotics branch of AI \citep{BILLARD98,Brooks2002,cangelosi2010roadmap,MIROLLI2011298}.

Another influential perspective on child development are Jean Piaget's foundational theories of cognitive development \citep{piaget1963}.
For Piaget, the child is a solitary thinker. While he acknowledged that social context can assist development, for him cognitive maturation happens mainly through the child's solitary exploration of their world. The child is a "little scientist" deciding which experiments to perform to challenge its assumptions and improve its representation of the world.

This Piagetian view on development is well aligned with mainstream Deep Reinforcement Learning research, which mainly focuses on sensorimotor development, through navigation and object manipulation problems rather than language based social interactions \citep{dqn,ddpg,her}. On the other hand, the study of language has been mostly separated from DRL, into the field of Natural Language Processing (NLP), which is mainly focused on learning (disembodied) language models for text comprehension and/or generation (e.g. using large text corpora as in \citet{gpt3}).

In the last few years however, recent advances in both DRL and NLP made the Machine Learning community reconsider experiments with language based interactions \citep{survey-rl-nlp,bender-2020-climbing-NLU}. Text-based exploratory games have been leveraged to study the capacities of autonomous agents to properly navigate through language in abstract worlds \citep{textworld,fantasy-text-world-2020,how-to-dragon-2020}. While these environments allow meaningful abstractions, they neglect the importance of embodiment for language learning, which has long been identified as an essential component for proper language understanding and grounding \citep{cangelosi2010roadmap,Bisk_2020}. Following this view, many works attempted to use DRL to train embodied agents to leverage language, often in the form of language-guided RL agents \citep{chevalierboisvert2018babyai,colas-imagine, Hill2020instrfollow,akakzia2021} and Embodied visual Question Answering (EQA) \citep{eqa,DBLP:conf/cvpr/GordonKRRFF18}, and more recently on interactive question production and answering \citep{imitating-intel}. Language use has also been studied in Multi-agent emergent communication settings, both in embodied and disembodied scenarios \citep{DBLP:conf/aaai/MordatchA18,jacques2019-socialinfl,DBLP:conf/iclr/Lowe0FKP20,DBLP:conf/aaai/WoodwardFH20}.

One criticism that could be made over aforementioned works in light of Vygotsky's theory is the simplicity of the "social interactions" and language-use situations that are considered: in language-conditioned works, the interaction is merely just the agent receiving its goal as natural language within a simple and rigid interaction protocol \citep{survey-rl-nlp}. In EQA, language-conditioned agents only need to first navigate and then produce simple one or two words answers. And because of the complexity of multi-agent training, studies on emergent communication mostly consider simplistic language (e.g. communication bits) and tasks. 

To catalyze research on building proficient social agents, we propose to identify a richer set of socio-cognitive skills than those currently considered in most of the DRL and NLP literature. We organize this set along 3 dimensions. Proficient social agents must be able to master \textit{intertwined multimodality}, i.e. coordinating multimodal actions based on multimodal observations. They should also be able to build an (explicit or implicit) \textit{theory of mind}, i.e. inferring other's mental state, e.g. beliefs, desires, emotions, etc. Lastly, they should be able to learn diverse and complex \textit{pragmatic frames}, i.e. social interaction protocols described as "verbal or non-verbal patterns of goal-oriented behaviors that evolve over repeated interactions between learners and teachers" \cite{bruner85pragframes}. 

Based on these target socio-cognitive skills, we present \benchname, a set of grid-world environments as a first step to foster research in this direction (see fig. \ref{fig:all_envs}). To study complex social scenarios in reasonable computational time, we consider single-agent learning among scripted agents (a.k.a. Non-Player-Characters or NPCs) and use low-dimensional observation and action spaces. We also use templated language, enabling to emphasize the under-studied challenges of dealing with more complex and diverse social and pragmatic situations. To showcase the relevance of \benchname, we study the failure case of a current SOTA DRL approach on this benchmark through detailed case studies.



\vspace{0.2cm}\textbf{Social agents are not objects.~~} Although social peers could be seen as merely complex interactive objects, we argue they are in essence quite different.
Social agents (e.g. humans) can have very complex and changing internal states, including intents, moods, knowledge states, preferences, emotions, etc.
The resulting set of possible interactions with peers (social affordances) is essentially different than those with objects (classical affordances). In cognitive science, an affordance refers to what things or events in the environment afford to an organism \citep{social_affordance}.
A flat surface can afford "walking-on" to an agent, while a peer can afford "obtaining directions from".
The latter is a social affordance, which may require a social system and conventions (e.g. politeness), implying that social peers have complex internal states and the ability to reciprocate. Successful interaction might also be conditioned on the peer's mood, requiring communication adjustments.

Training an agent for such social interactions most likely requires drastically different methods -- e.g. different architectural biases -- than classical object-manipulation training.
In \benchname~we simulate such social peers using scripted NPCs. We argue that studying isolated social scenarios featuring NPCs in tractable environments is a promising first step towards designing proficient social agents.

\vspace{0.2cm}\textbf{Grounding language in social interactions.}
In AI, \textit{natural language} often refers to the ability of an agent to use a large vocabulary and complex grammar. We argue that this is but one dimension of the \textit{naturalness} of language.
Another, often overlooked, dimension of this \textit{naturalness} refers to language grounding, i.e. the ability to anchor the meaning of language in physical, pragmatic and social situations \citep{steels2007symbolgrounding}.
The large literature on language grounding has so far mostly focused on grounding language into physical action \citep{cangelosi2010roadmap,chevalierboisvert2018babyai,colas-imagine}: here the meanings of sentences refer to actions to be made in interaction with objects (e.g. "Grasp the blue box"). However, natural language as used by humans is also strongly grounded in social contexts: not only the interpretation of language requires understanding the social context (e.g. taking into account intents or beliefs of others), but meanings can refer to social actions (e.g. "Help your friend to learn his dance lesson"). Here, an important aspect of language naturalness refers to the diversity of kinds of pragmatic social situations in which it is grounded: the work presented here aims at making steps in this direction.



\vspace{0.2cm}\textbf{Main contributions:}
\begin{itemize}
\item An outline of a set of core socio-cognitive skills necessary to enable artificial agents to efficiently act and learn in a social world.
\item \benchname, a set of grid-world environments including complex social situations with scripted NPCs to benchmark the capacity of DRL agents to learn socio-cognitive skills organized across several dimensions.
\item Performance assessment of a SOTA Deep RL approach on \benchname~and analysis of its failure using multiple case studies.
\end{itemize}




\section{Related Work}

An extended version of the following related work is available in appendix \ref{app:extended-related}.

\textbf{Earlier calls for socially proficient agents~~} This work aims to connect the recent DRL \& NLP literature to the older developmental robotics field \citep{asada2009-dev-rob,cangelosi2014dev-rob}, which studies how to leverage knowledge from the cognitive development of human babies into embodied robots. Within this field, multiple calls for developing the social intelligence of autonomous agents have already been formulated \citep{billard99,lindblom-vygot-and-beyond,MIROLLI2011298}. Vygotsky's emphasis on the importance of social interactions for learning is probably what led Bruner to conceptualize the notion of pragmatic frames \citep{bruner85pragframes}, which has later been reused to theorize language development \citep{pragmatic_frames_lang_acquisition}. In this work we intent to further motivate the relevance of this notion to catalyse progress in Deep RL and AI. 

\textbf{Human-Robot Interaction~~} Interactions with knowledgeable human teachers is a well studied form of social interaction. Many works within the Human-Robot Interaction (HRI) and the Interactive Learning field studied how to provide interactive teaching signals to their agents, e.g. providing instructions \citep{grizou-2014-hri-instructions}, demonstrations \citep{argall-LfD-2009}, or corrective advice \citep{celemin-advice}.  In \citep{pragmatic-frames}, authors review this field and aknowledge a lack of diversity and complexity in the set of studied pragmatic frames. Echoing this observation, the \benchname~benchmark invites to study a broader set of social situations, e.g. requiring agents to both move and speak, and even to learn to interact in a \textit{diversity} of pragmatic frames. Catalysing research on DRL and social skills seems even more relevant now that many application-oriented works are beginning to leverage RL and DRL into real world humanoid social robots \citep{socialRobots2021}.

\textbf{Recent works on language grounded DRL~~} Building on NLP, developmental robotics, and goal-conditioned DRL \citep{colas-goal-cond-2020}, many recent works presented embodied DRL-based agents able to process language \citep{survey-rl-nlp}. Most of this research concerns language conditioned agents in \textit{instruction following} scenarios (e.g. "go to the red bed"), and present various solutions to speed-up learning, such as auxiliary tasks \citet{Hermann-grounded-lang-2017}, pre-trained language models \citep{hill2020b-instr-fol-transfer}, demonstrations \citep{fu2018from,Lynch2020ground-lang-in-play}, or descriptive feedbacks \citep{colas-imagine,Nguyen-activity-description-2021}. In Embodied Visual Question Answering works, agents are conditioned on questions, requiring them to navigate within environments and then produce an answer ("what color is the bed ?") \citep{eqa,DBLP:conf/cvpr/GordonKRRFF18}. Compared to these works, \benchname~aims to enlarge the set of considered scenarios by studying how to ground and produce language within diverse forms of social interactions among embodied social peers.

Closely related to our work, \cite{imitating-intel}
present experiments on a simulated 3D environment designed for multi-agent interactive scenarios between embodied multi-modal agents and human players. Authors propose novel ways to leverage human demonstrations to bootstrap learning.
Because of the complexity of their setup (imitation learning, 3D environments, pixel-based, human in the loop, ...), only two now-common social interaction scenarios are studied: question answering (i.e. EQA) and instruction following. The novelty of their work is that these questions/instructions are alternatively produced and tackled by learning agents. \benchname~focuses on a lighter experimental pipeline (2D grid-world, symbolic pixels, no humans) to enable the study of a broader range of social scenarios, requiring multi-steps conversations and interactions with multiple (scripted) agents.

\textbf{Benchmarks on embodied agents and language~~}
Benchmarks featuring language and agents embodied in physical worlds already exists, however many of them only consider the aforementioned instruction-following \citep{chevalierboisvert2018babyai,MisraBBNSA18,RuisBench2020} and question-answering \citep{eqa,DBLP:conf/cvpr/GordonKRRFF18} scenarios.
In between disembodied NLP testbeds \citep{wang-etal-2018-glue,social-IQ-2019} and previous embodied benchmarks is the LIGHT environment \citep{UrbanekLIGHT}, a multiplayer text adventure game allowing to study social settings requiring complex dialogue production \citep{how-to-dragon-2020,fantasy-text-world-2020}. Instead of the virtual embodiment of text-worlds, \benchname~tackles the arguably harder and richer setting of egocentric embodiment among embodied social peers.
Within the Multi-Agent RL field, \cite{DBLP:conf/aaai/MordatchA18} propose embodied environments to study the emergence of grounded compositional language. Here language is merely a discrete set of abstract symbols used one at a time (per step). While symbol negociation is an interesting social situation to study, we leave it to future work and consider scenarios in which agents must enter an already existing social world (using non-trivial language). In \cite{jacques2019-socialinfl}, authors present Multi-Agent social dilemna environments requiring the emergence of cooperative behaviors through \textit{non-verbal} communication. We consider both non-verbal (e.g. gaze following) and language-based communication.

\section{Social skills for socially competent agents}
\label{sec:soc-skills}

Social skills have been extensively studied in cognitive science \citep{social-skills-assessment1986,social-skills-framework2010} and developmental robotics \citep{cangelosi2010roadmap}. Based on these works, this section identifies a set of core social skills when aiming to train  socially competent artificial agents.


\vspace{0.2cm}\textit{\textbf{1 - Intertwinded multimodality~~}} refers to the ability to interact and use multiple modalities (verbal and non-verbal) in a coordinated manner. A proficient agent should be able to act using both primitive actions (moving) and language actions (speaking), and to process both visual and language observations of social peers.
Importantly, socially competent agents must be able to interact using multiple modalities in an \textit{intertwined} fashion.
By intertwined multimodality we refer to the ability to adapt its multimodal interaction sequence, rather than following a pre-established progression of modalities.
For example, in EQA\citep{eqa}, the progression is always as follows: 1) a question is given to the agent at the beginning of the episode, 2) the agent moves through the environment to gather information, and 3) upon finding an answer it responds (in language) and the episode ends.
By the term \textit{intertwined multimodality} we aim to emphasize that the modalities often interchange and the question of "when to use which modality" is non-trivial, e.g. sometimes the relevant information can be obtained by \textit{asking} for it and sometimes by \textit{looking} for it.

\vspace{0.2cm}\textit{\textbf{2 - Theory of Mind(ToM)~~}} refers to the ability of an agent to attribute to others and itself mental states, including beliefs, intents, desires, emotions and knowledge \citep{wellman1992childToM,flavell1999ToM}.
An agent that has ToM perceives other participants as \textit{minds} like itself.
This enables the agent to theorise about other's intents, knowledge, lack of knowledge etc.
Here we outline some, of many, different perspectives of ToM to better demonstrate how ToM is essential for human social interactions:
    
    \textbf{Inferring intents:} the agent is able to infer, based on verbal or non-verbal cues, what others will do or want to do, e.g. that some social peers are liars/trustworthy.
    
    \textbf{False belief:} the agent understands that someone's belief (including its own) can be faulty \citep{bailla2010falsebeliefinfants}.
    
    \textbf{Imitating or emulating social peer's behaviour:} agent can imitate a behaviour or a goal seen in a social peer, e.g. upon observing a peer cut onions the agent is able to cut the onions himself, either with the same movement or with its own strategy.

    \vspace{0.2cm}\textit{\textbf{ 3 - Pragmatic frames ~~}}
    refer to the regular patterns characterizing the unfolding of possible social interactions (equivalent to an interaction protocol or a grammar of social interactions).
    Pragmatic frames simplify learning by providing a stable structure to social interactions.
    An example of a pragmatic frame are the turn taking games.
    By playing those games a child extracts the rule of each participant having his "turn". It can then generalize this rule to a conversation where it understands that it shouldn't speak while someone else is speaking. We propose to outline several facets of pragmatic frames that proficient social agents should be able to master:
    
    \textbf{Learning a pragmatic frame} The agent is able to learn a frame through social interactions, without it being manually hand-coded (e.g. as in instruction-following scenarios). Through rich social interactions (e.g. dialogues) with one or several peers, the agent should be able to infer the structure of the interaction pattern (frame), extract potential instructions, and leverage them appropriately.
    
    \textbf{Teaching frames }
    are a specific type of pragmatic frames involving a teacher explicitly teaching a certain content via a \textit{slot}. A slot refers to the place in the interaction sequence holding the variable learning content.
    A parent teaching a child words with the help of a picture book is one such teaching frame. Upon seeing a picture with a dog a parent might point to the dog, say "Look, it's a dog", and establish eye contact to verify that the child understood the message.
    Upon, however, seeing a picture of a cat he might say "Look, it's a cat".
    Here "dog" and "cat" are learning contents and the slot is the location of those words in the sequence ("Look, it's a \textit{<slot>}").
    A socially competent agent should be able to learn such a frame and extract the \textit{learning content} from it.
    
    \textbf{Roles - } An agent is able to not only understand the relevance of various participants for achieving a shared goal but also learn about the others' role just from playing its own.
    For example, in a setting where one agent opens the door to enable another agent to exit the room, the \textit{exiting} agent should be able to learn what the role of opening the door consists of. This exiting agent should then be able to open the door for another agent with little or no additional training. The social interaction described above consists of one frame viewed from two different perspectives corresponding to two different roles. Socially proficient agents should be able to easily learn this whole frame by experiencing it just from their own perspective.

    \textbf{Diversity - }
            the agent can learn many different frames and differentiate between them. Furthermore, the agent can reuse those frames in new situations and even negotiate and construct new ones.
            
    \textbf{Frame changes - } an agent is able to detect and adjust to a change of the current pragmatic frame. For example, while playing football we are able to participate in \textit{small talk} with another player.

\section{The SocialAI 1.0 Benchmark}
\label{sec:socialai_benchmark}
\begin{figure*}[t!]
\centering
\subfloat[TalkItOut]{\includegraphics[width=0.23\textwidth]{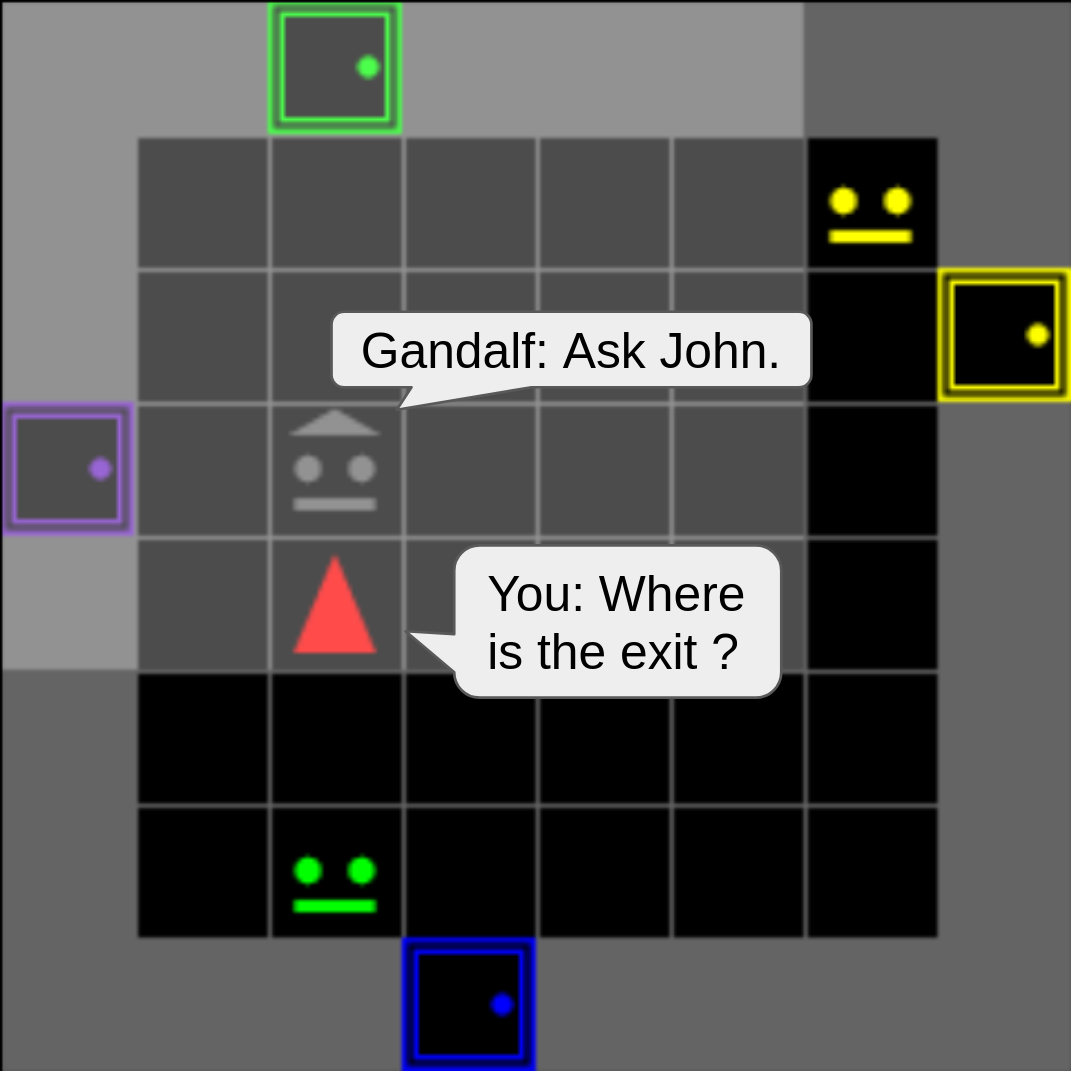}}
\subfloat[Dance]{\includegraphics[width=0.23\textwidth]{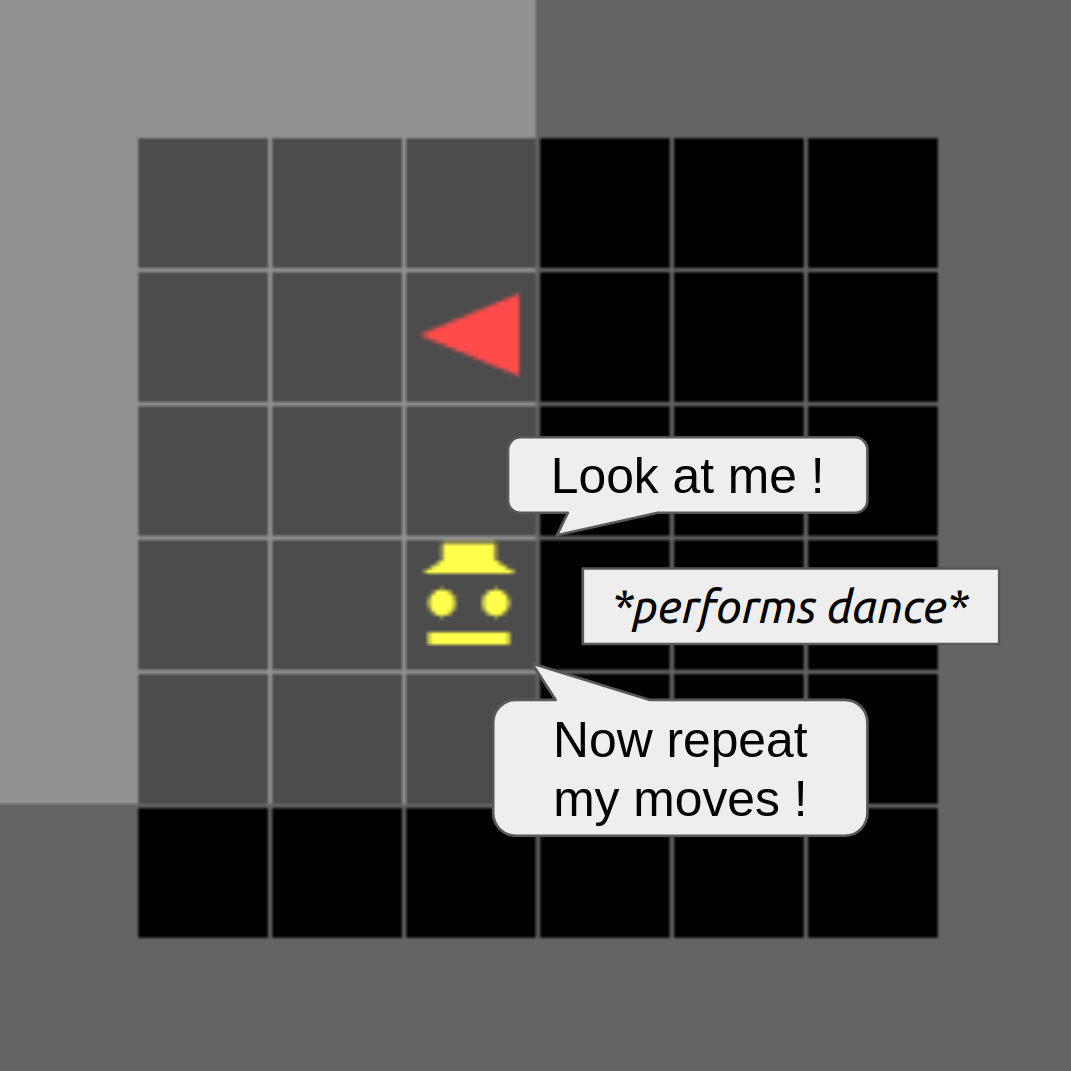}}
\subfloat[CoinThief]{\includegraphics[width=0.23\textwidth]{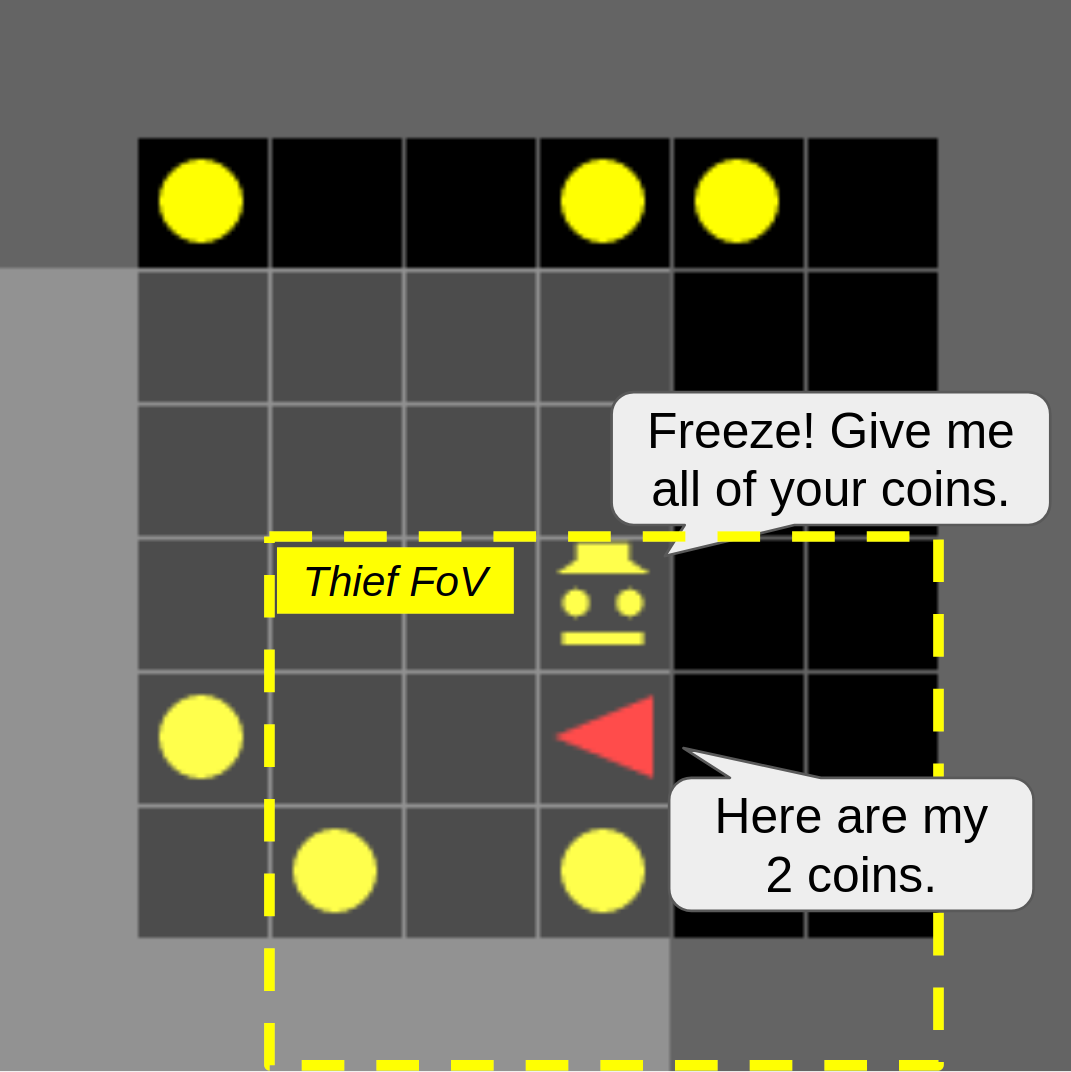}}

\subfloat[DiverseExit]{\includegraphics[width=0.23\textwidth]{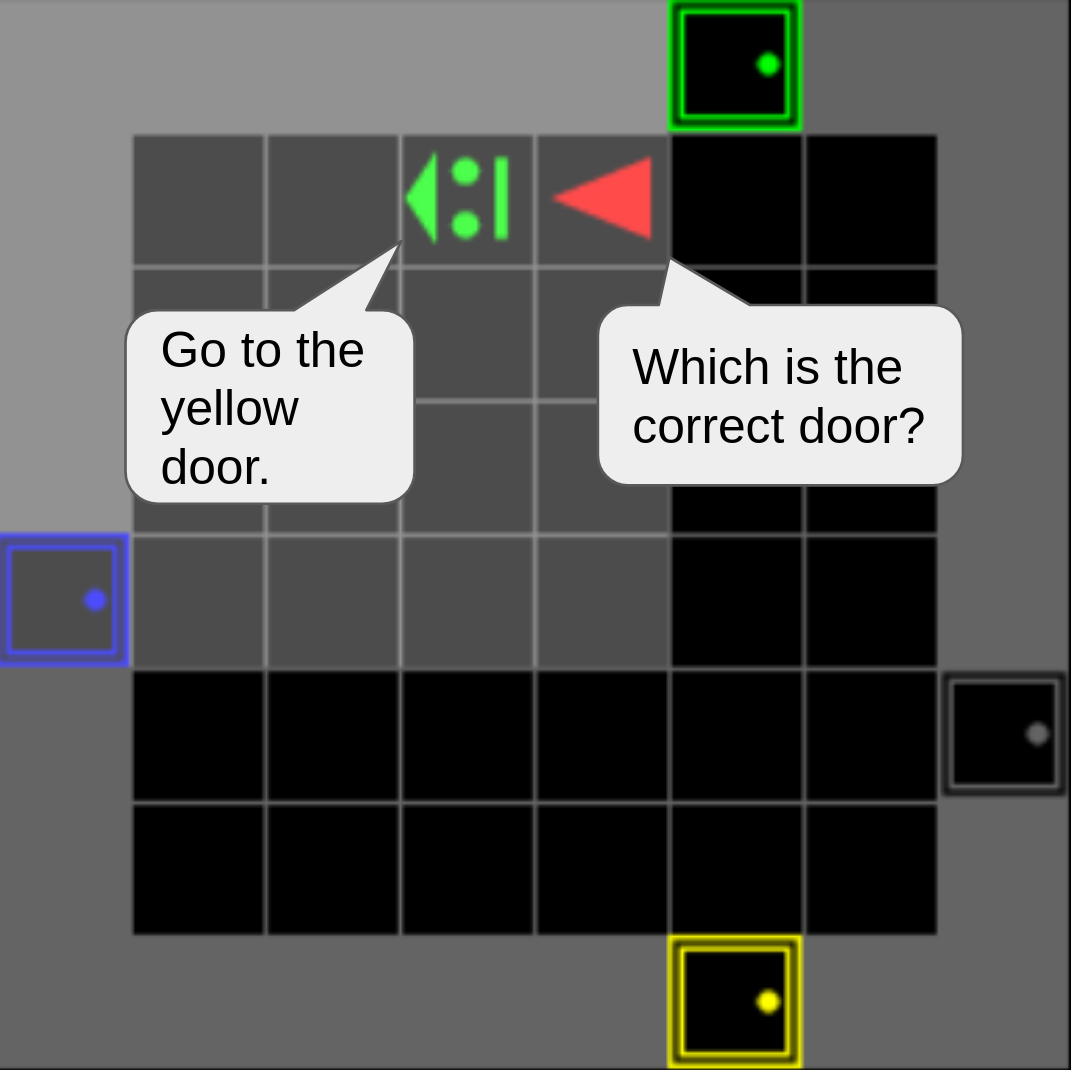}}
\subfloat[ShowMe]{\includegraphics[width=0.23\textwidth]{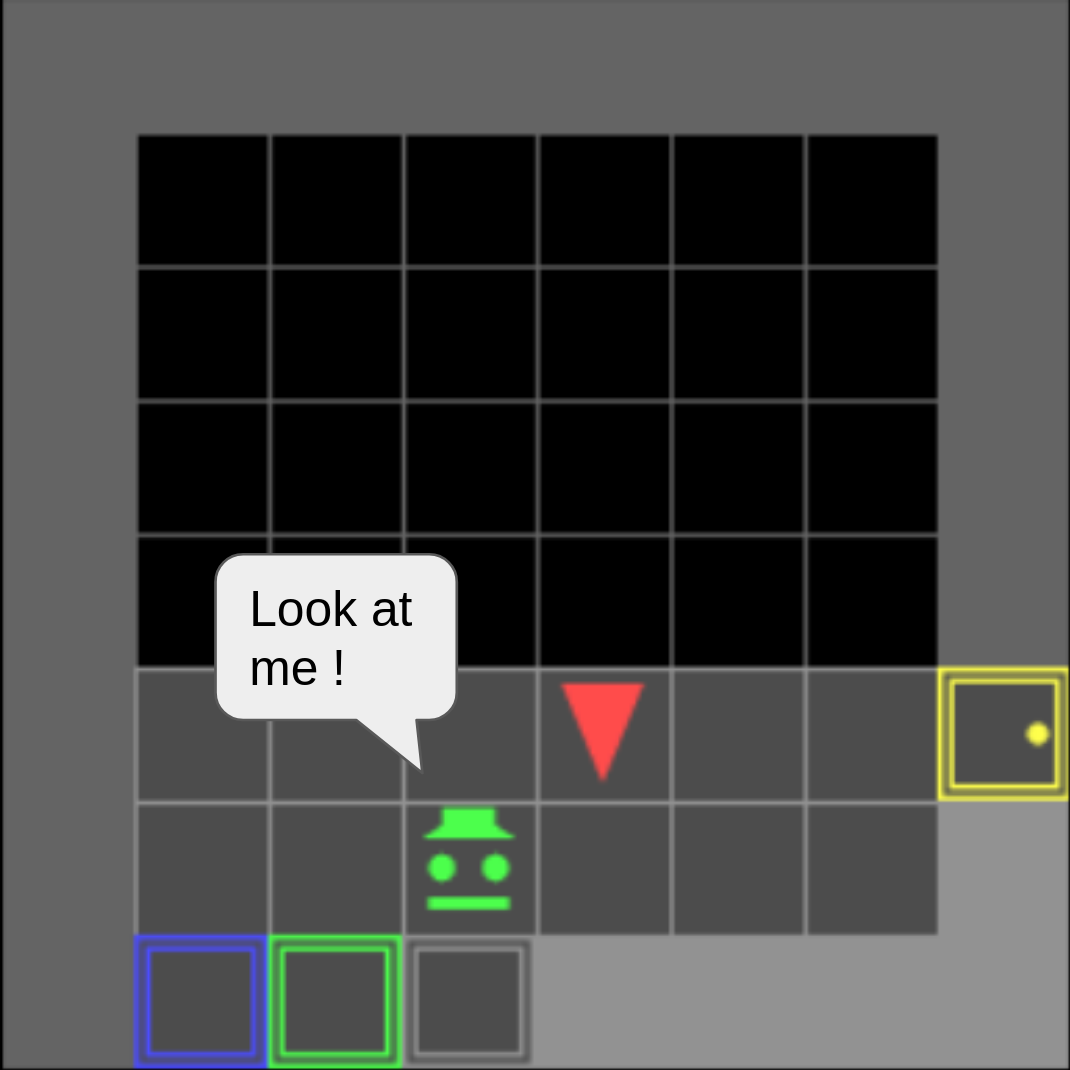}}
\subfloat[Help]{\includegraphics[width=0.23\textwidth]{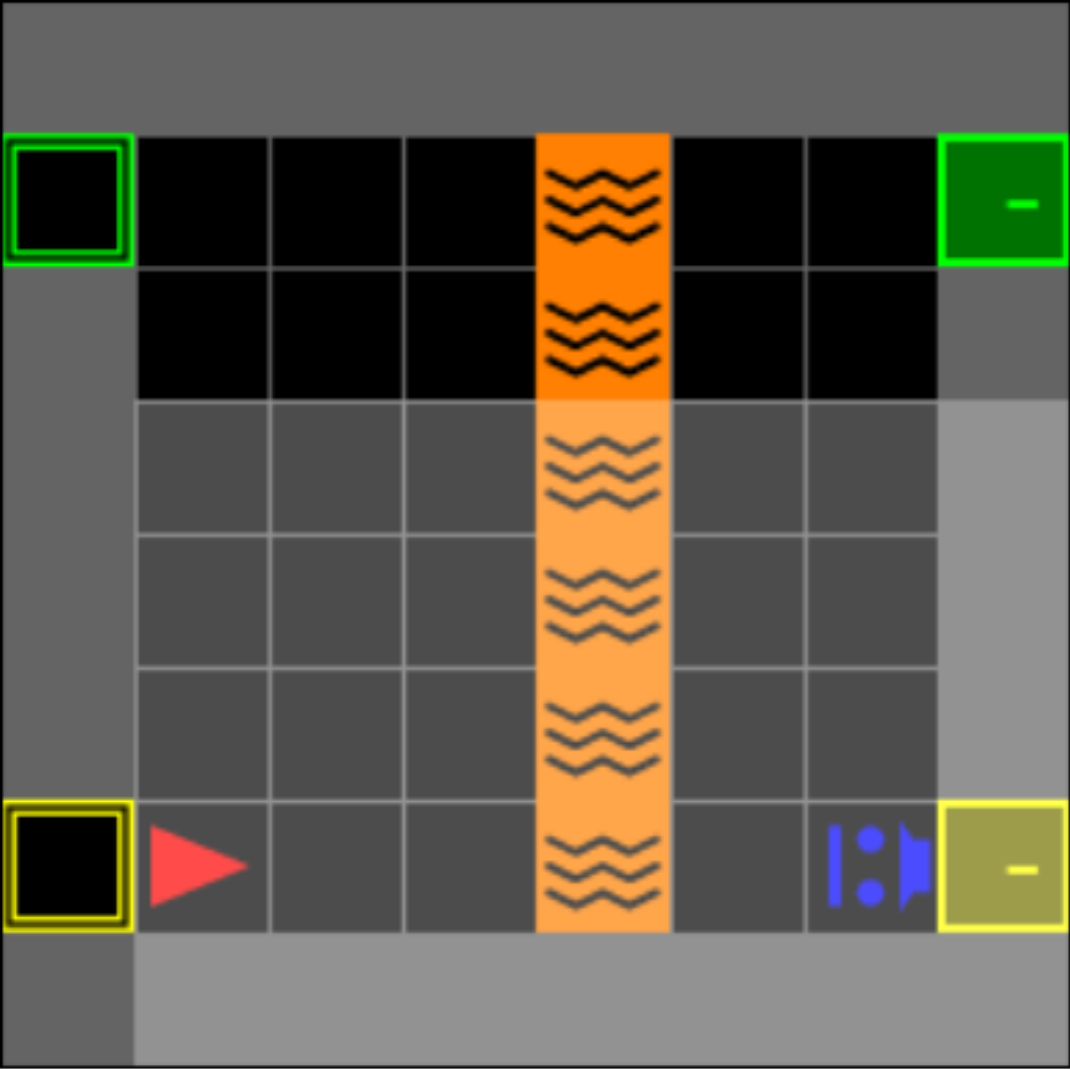}}
\subfloat[Legend]{\includegraphics[width=0.15\textwidth]{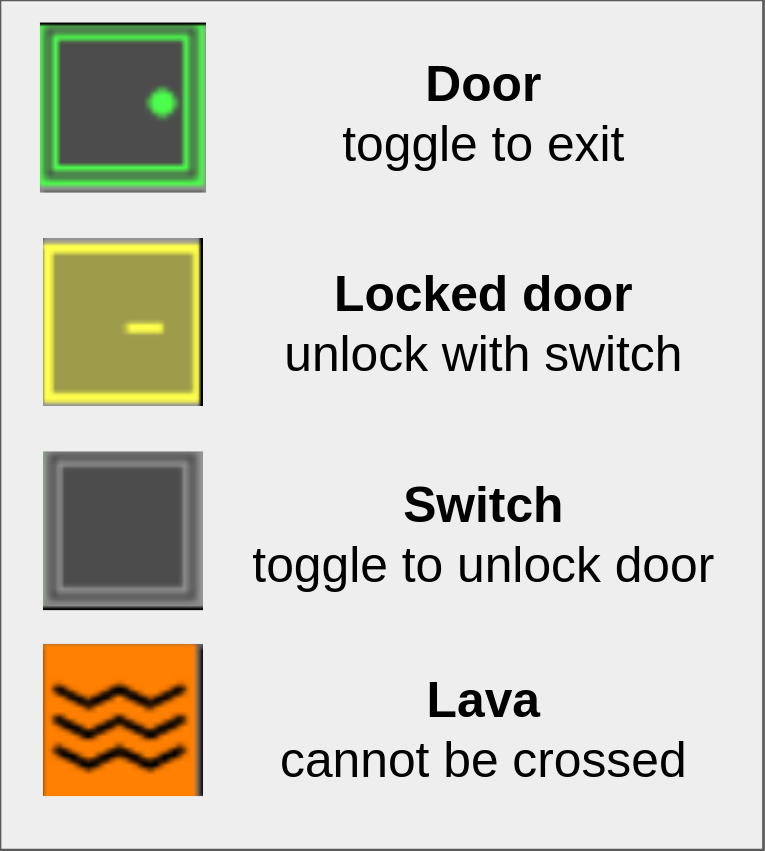}}

\caption{\footnotesize{\textbf{SocialAI 1.0} is composed of multiple grid-world environments featuring scripted NPCs. Solving this benchmark requires to train socially proficient Deep Reinforcement Learning agents.}}
\label{fig:all_envs}
\end{figure*}

To catalyse research on developing socially proficient autonomous agents, we present SocialAI 1.0 (see fig. \ref{fig:all_envs}), a set of grid-world environments designed to challenge learners on important aspects of the core social skills mentioned in sec. \ref{sec:soc-skills}. In this section we briefly present each environment (see app. \ref{app:env-details} for details) and highlight how they require various subsets of the aforementioned core social skills.

\paragraph{Common components} The key design principle of \benchname~environments is to allow the study of complex social situations in reasonable computational time. As such, we consider single-room grid-world environments (8x8 grid), based on minigrid \citep{gym_minigrid}(Apache License). The learning agent can both navigate using discrete actions (e.g. turn left/right, go forward, toggle) and use template-based language generation (environment-dependent). As observations, the agent receives a partial 7x7 agent-centric symbolic pixel grid (see highlighted cells in fig. \ref{fig:all_envs}), with 4 dimensions per cell (type, color, status, orientation). Additionally, the agent receives the history of observed language outputs from NPCs preceded by the NPC's name (ex. "John: go to the red door). A positive reward is given only upon successful completion of the social scenario (discounted by time taken).

In the following description of environments, unless stated otherwise, the agent, all objects and all NPCs are spawned randomly for each new episode. Each description highlights the socio-cognitive skills required to solve the environment (see table \ref{tab:env_skills} for a recapitulating overview).

    \textbf{\textit{TalkItOut -}} The agent has to exit the room using one of the four doors (by uttering "Open Sesame" in front of it). The environment features a wizard and two guides (one lying, one trustworthy). To find out which door is the correct one the agent has to ask the trustworthy guide for directions, and to find out which guide is trustworthy it has to query the wizard (which requires a preliminary politeness formula: "Hello, how are you?"). Solving TalkItOut requires mastering \textit{intertwined  multimodality}, basic \textit{Theory of Mind} (infering ill-intentions), and a basic \textit{pragmatic frame} (the agent must stand near NPCs to interact with them).
    
    \textbf{\textit{Dance -}} A NPC demonstrates a 3-steps dance pattern (randomly generated for each episode) and then asks the agent to reproduce this dance. Each dance step is composed of a movement action and, half of the time, of an utterance. To solve \textit{Dance}, agent must reproduce the full dance sequence. Multiple trials are authorized. Only trials performed after the NPC completed his dance are recorded. This requires the agent to be able to infer that the NPC is setting up a \textit{teaching pragmatic frame} ("Look at me" + \texttt{do\_dance\_performance} + "Now repeat my moves"), requiring the agent to \textit{imitate} a social peer, process \textit{multi-modal observations} and produce \textit{multi-modal actions}.
    
    \textbf{\textit{CoinThief -}} In a room containing 6 coins, a thief NPC spawns near the agent, and utters that the agent must give "all of its coins". To obtain a positive reward, the agent must give (using language) exactly the number of coins \textit{that the thief can see} (the thief field of view is a 5x5 square, i.e. a smaller version than the agent's). This requires \textit{Theory of Mind} as the agent must understand that the thief holds \textit{false belief} over the agent's total number of coins and must infer how many coins he actually sees.
    
    
    \textbf{\textit{ShowMe -}} The agent has to exit the room through the locked door. To unlock the door it has to press the correct button, and to find out which button is the correct one it has to look at the NPC. The NPC waits for the agent to establish eye contact, then presses the correct button and exits the room. Solving ShowMe requires that the agent infers the \textit{teaching pragmatic frame} and imitates the NPC's goals (pressing a button, and exiting the room).
    
    \textbf{\textit{DiverseExit -}} The agent has to exit the room using the correct door (one out of four). To find out which door is the correct one it has to ask the NPC. There are twelve different NPCs which can be present in the environment (each episode a random one is chosen).
    Each NPC prefers to be asked (using language) for directions differently (e.g. by standing close, by poking him, etc), i.e. via a different pragmatic frame.
    To solve DiverseExit the agent has to learn the diversity of frames and, most importantly, which one to use with which NPC. 
    
    \textbf{\textit{Help -}} the environment consists of two roles (the Exiter and the Helper), one played by the agent and another by the NPC. The Exiter is placed on the right side of the room and has to exit the room using one of the two doors 
    The doors are locked and each has a corresponding unlocking switch on the left wall.
    The episode ends without reward if both switches are pressed.
    The Helper, placed on the left side of the room, has to press the switch unlocking the door by which the agent wants to exit. 
    The agent is trained in the Exiter role, but tested in the Helper role. To solve Help the agent needs to learn about both roles just from training as the Exiter. i.e. learn the full pragmatic frame just from seeing its own perspective of it.
    
    \textbf{\textit{SocialEnv -}} In this meta-environment, which contains all previous ones, we consider a multi-task setup, in which the agent is facing a randomly drawn environment, i.e. it has to infer what is the current social scenario he is spawned in (using pragmatic information collected through interaction). Mastering this environment requires to be proficient in \textit{all of the core social skills} we proposed. 

\begin{table}[htb!]
\caption{List of core socio-cognitive skills required in each environment.}
\label{tab:env_skills}
\begin{center}
\begin{tabular}{@{}r|cccccccc@{}}

Social Skills \textbackslash SocialAI Envs      & \rot{TalkItOut} & \rot{Dance} & \rot{CoinThief} & \rot{DiverseExit}  & \rot{ShowMe} & \rot{Help} & \rot{SocialEnv} \\ \midrule
Intertwined m.m.                    & $++$              & $++$      & $+$       & $++$          & $++$          &          & $++$               \\
\hline
ToM - inferring intent      & $++$              & $+$       & $+$       & $++$          & $+$           & $++$      & $++$               \\
ToM - false belief          &                   &           & $++$      &               &               &          & $++$               \\
ToM - imitating peers &                   & $++$      & $+$          &               &               & $+$      & $++$               \\
ToM-joint attention         &                   &           &           & $++$          & $+$           & $+$      & $++$               \\
\hline
P. Frames - Diversity       & $+$               & $+$       & $+$       & $+$           & $++$          & $+$      & $++$               \\
P. Frames -Teaching         &                   & $+$       &           &               & $+$           &          & $++$               \\
P. Frames - Roles           &                   &           &           &               &               & $++$     & $++$               \\ \bottomrule
\end{tabular}
\end{center}
\end{table}

\section{Experiments and Results}

To showcase the relevance of \benchname~as a testbed to assess the socio-cognitive abilities of DRL agents, and to provide initial target baselines to outperform, we test a recent DRL architecture on our environments. Through global performance assessment and multiple case-studies, we demonstrate that this agent essentially fails to learn due to the \textit{social complexity} of \benchname's environments.


\paragraph{Baselines} 
Our main baseline is a PPO-trained \citep{ppo} DRL architecture proposed in \cite{hui2020babyai}. We chose this model as it was designed for language-conditioned navigation in grid worlds, which is similar to our setup (although in our case language input is not fixed but varies along interactions). We modify the original architecture to be Multi-Headed, since our agent has to both navigate and talk, and thereafter name the resulting condition \textit{PPO}.
We also consider a variation of this baseline trained with additional intrinsic exploration bonuses (\textit{PPO+Explo}). We consider two different exploration bonuses to reward the discovery of either new utterances or new visual objects. In each environments, we determined empirically the optimal set of exploration bonus (visual only, utterance only, or both), and only report results for the best configuration. Finally, as a lower-baseline, we consider an ablated, non-social version of our PPO agent, from which observation inputs emanating from NPCs are removed (\textit{Unsocial PPO}). See appendix \ref{app:cond-details} for details.

\paragraph{Overall results} For each condition, 16 seeded runs of $30$ Millions environment steps are performed on each environments. Performance is defined as the percentage of episodes that were solved (success rate). Post-training performance results are gathered in Table \ref{tab:results}. All considered agents essentially fail to learn, on all environments.
In \textit{ShowMe}, \textit{DiverseExit} and \textit{Help}, both PPO and PPO with exploration bonus (PPO+Explo) performance are not statistically significantly different from Unsocial PPO, our lower-baseline agent that doesn't observe the NPC ($p > 0.5$ in all cases, using a post-training Welch's t-test). This implies that our agents are not able to leverage NPC-related inputs, i.e. they are not socially proficient.
On both \textit{TalkItOut} and \textit{DiverseExit}, PPO agents converge to a local optimum of $25\%$ success rate, which corresponds to ignoring the NPC and going to any door. 


To better understand why our agents are failing to learn (and as a sanity check for our implementations), we present additional performance analysis of three environment-specific case studies highlighting different social skills categories of sec. \ref{sec:soc-skills}: \textit{TalkItOut} (Intertwined Multimodality), \textit{CoinThief} (Theory  of Mind), and \textit{Help} (Pragmatic Frames).




\begin{table}[]
\caption{Success rates (mean $\pm$ std. dev.)  of considered baselines on \benchname~after $30$ Millions environment steps (on a fixed test set of 500 environments). Our DRL agents fails to learn.}
\vspace{0.4cm}
\centering
\begin{tabular}{@{}llll@{}}
\toprule
Env \textbackslash Cond  & PPO                  & PPO + Explo           & Unsocial PPO       \\ \midrule
TalkItOut               & $0.25 \pm 0.01$     & $ 0.12 \pm 0.03$    & $0.25 \pm 0.01 $    \\
Dance                   & $0.03 \pm 0.01$       & $0.03 \pm 0.01$       & $0.01 \pm 0.0$   \\
CoinThief               & $0.45 \pm 0.08$       & $0.47 \pm 0.04$       & $0.38 \pm 0.02$   \\
DiverseExit             & $0.25 \pm 0.02$     & $0.25 \pm 0.01$    & $0.24 \pm 0.01 $   \\
ShowMe                  & $0.0 \pm 0.0$         & $0.0 \pm 0.0$         & $0.0 \pm 0.0 $     \\
Help                    & $0.12 \pm 0.05$     & $0.11 \pm 0.04$     & $ 0.15 \pm 0.06 $     \\
SocialEnv               & $0.06 \pm 0.02$     & $0.08 \pm 0.02$     & N/A   \\ \bottomrule
\end{tabular}
\label{tab:results}

\end{table}


\paragraph{Case study - TalkItOut} \textit{TalkItOut} is challenging because the agent has to master a non-trivial progression of modalities. To talk with an NPC, apart from the language modality, both vision, and primitive actions have to be used to move close to the NPC (which is mandatory for communication). 
Furthermore, the agent has to learn to infer, from verbal-cues of the dialogue with the wizard, which guide is the ill-intended one (i.e. a facet of ToM). For this experiment we construct an ablation environment where the ill-intended NPC is removed, which greatly reduces the social complexity as 1) all NPCs are now well-intended, and 2) dialogue with only the trustworthy guide is sufficient to solve the task.

\begin{figure}[htb!]
\centering
\subfloat[TalkItOut]{\includegraphics[width=0.32\textwidth]{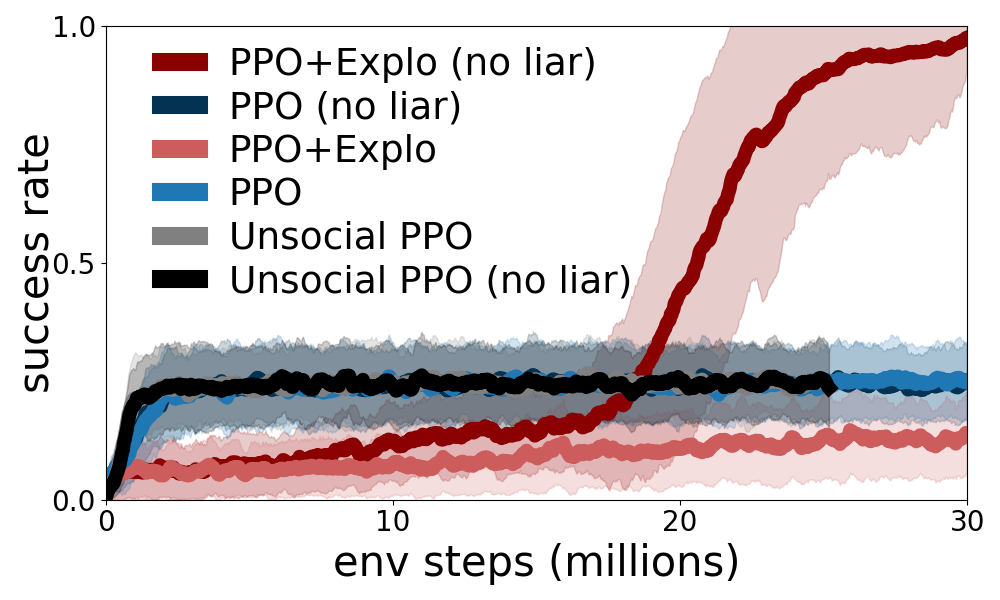}}
\subfloat[CoinThief]{\includegraphics[width=0.32\textwidth]{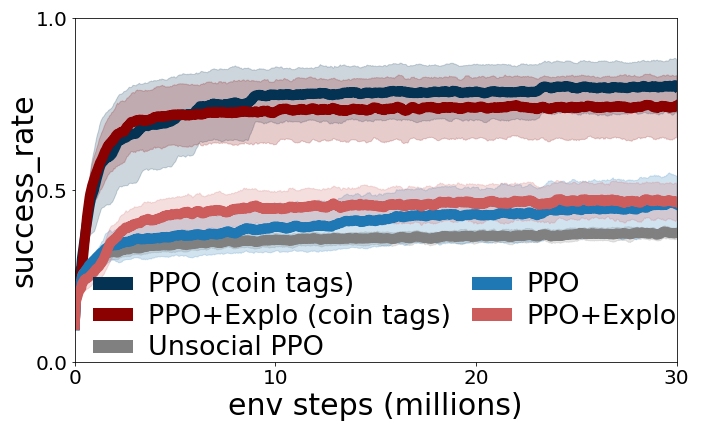}}
\subfloat[Help]{\includegraphics[width=0.32\textwidth]{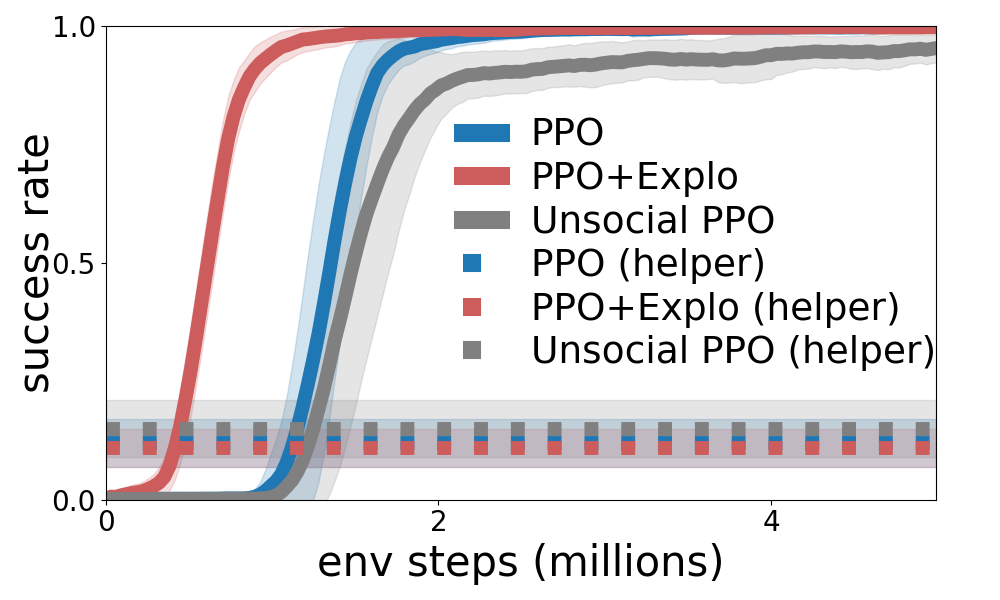}}
\caption{Evolution of success rates along training in three environment specific case-studies. Mean and std. deviation plotted, 16 seeds per condition.}
\label{fig:case-studies-results}
\end{figure}

\textit{Results - } Figure \ref{fig:case-studies-results}a shows the training success rates of all our baselines.
We can see that, in both environments, the PPO condition gets stuck at $25\%$ success rate, i.e. the local optimum of ignoring the NPC and going to a random door. Adding exploration bonus (PPO+Explo condition) enables the agent to overcome this local optimum, however only in the ablation environment does this result in solving the task. This shows that the social complexity introduced by the lying guide is too challenging for our conditions. These experiments suggest that our agents lack sufficient biases for both mastering intertwined multimodal interactions and inferring different intents of social peers.

\paragraph{Case-study - CoinThief}To assess whether it is the social complexity of the CoinThief environment that prevents our agents to learn high-performing policies, we consider a simplified version of the environment in which coins visible to the NPC have a different visual encoding from other coins. This modification removes the need to infer the NPC's field of view, i.e. the correct number of coins can be given to the NPC without any form of social awareness.

\textit{Results -} Performance curves for our PPO variants on CoinThief and on the simplified CoinThief (with \textit{coin tags} for NPC-visible coins) are shown in figure \ref{fig:case-studies-results}b. For both PPO and PPO+Explo, statistically significant improvements are obtained on the simplified environment w.r.t. vanilla CoinThief ($p<0.001$): both approaches respectively reach a final performance of $0.81$ and $0.75$ (not statistically significantly different, $p=0.07$).

\paragraph{Case study - Help}

The Help environment aims to test the ability of the agent to learn about the other's role from training only on its own i.e. to learn the whole pragmatic frame just from seeing its own perspective on it. We train the agent to achieve a shared goal on one role and then evaluate in a zero-shot manner on the other.





\textit{Results - } Figure \ref{fig:case-studies-results}c shows the training success rates for the agent in the Exiter role. The horizontal dotted lines depict the performance of the same final agents on the Helper role (depicted by the same colors).
We can see that in training the Exiter role is easily solved, reaching almost perfect success rate in less than two million environment steps. Furthermore, we can see that the agent with the exploration bonus (PPO+Explo) is able to solve the task faster.
When the same agents are evaluated in the Helper role the performance drastically drops ($\leq 15\%$ success rate). 
Qualitative analysis shows that this non-zero success rate on Helper role is due to agents acting as if in the Exiter role, which sometimes make them press the switch due to the stochastic nature of the PPO action sampling. The agent doesn't show any implication of understanding that the roles have been reversed.
These unsurprising results outline the inability of standard RL techniques to transfer the knowledge about the task to the opposite role. The agent only learns its perspective of the pragmatic frame and not the frame itself. It doesn't understand that its goal is shared with the NPC.


\section{Conclusion And Discussion}

In this work we classified and described a first set of core socio-cognitive skills needed to obtain socially proficient autonomous agents. We then presented \benchname -- an open-source testbed to assess the social skills of DRL learners, which leverages the computational simplicity of grid-world environments to enable the study of complex social situations. We then studied how a current SOTA DRL approach was unable to solve \benchname. By analyzing the failure cases of this approach through multiple case studies, we were able to highlight the relevance of our benchmark as a tool to catalyze future research on socially proficient DRL agents.

\textbf{A need for architectural biases.}
This work suggests that architectural improvements are needed for DRL agents to learn to behave appropriately in multimodal social environments. One avenue towards this is to endow agents with mechanisms enabling to learn models of others' minds, which has been identified in cognitive neuroscience works as a key ingredient of human social proficiency \citep{critiqueRLsociallearning2020}. Some ideas have already been formulated regarding how to enable agents to master theory of mind, such as by using a meta-learning approach (through the observation and modeling of populations of agents) \citep{DBLP:conf/icml/RabinowitzPSZEB18}, or by leveraging inverse RL \citep{jara2019-tom-irl}. This also points to the general open-question of what parts of biases need to be "innate", and what others could be learned through practicing diverse social interaction games in the lifetime of an agent. 

\textbf{Limitations.} What we present in this work is \benchname~\textit{version 1.0}, i.e. we expect this benchmark to evolve along the development of better learning architectures for agents learning in social worlds. As such, multiple interesting improvements over the current version of the benchmark could be considered. We could design NPCs with more elaborated internal states, e.g. by making them more adaptive to the learner's behavior. While we consider environments with fixed sets of pragmatic frames, another interesting avenue is to design environments with \textit{emergent} pragmatic frames, i.e. pragmatic frames that are negotiated between participants (a crucial component lacking from Human Robot Interaction methods \citep{pragmatic-frames}).


\textbf{Broader impact} Decision-making Machine Learning systems are more and more present in our everyday lives. In this work, we propose a fundamental research to catalyze the development of autonomous agents able to properly understand and act in a social world. Ultimately, this has the potential to simplify the alignment of machine behaviors to our human needs by easing communication.

\section*{Acknowledgments}
This work was supported by Microsoft Research through its PhD Scholarship Programme. All presented experiments were carried out using both A) the computing facilities MCIA (Mésocentre de Calcul Intensif Aquitain) of the Université de Bordeaux and of the Université de Pau et des Pays de l'Adour, and B) the HPC resources of IDRIS under the allocation 2020-[A0091011996] made by GENCI.

\bibliography{socialai}
\bibliographystyle{unsrtnat}

\clearpage

\appendix
\section{Supplementary Material: Extended Related Work}
\label{app:extended-related}

\paragraph{Earlier calls for socially proficient agents} This work aims to connect the recent DRL \& NLP literature to the older developmental robotics field \citep{asada2009-dev-rob,cangelosi2014dev-rob}, which studies how to leverage knowledge from the cognitive development of human babies into embodied robots. Within this field, multiple calls for developing the social intelligence of autonomous agents have already been formulated \citep{billard99,lindblom-vygot-and-beyond,MIROLLI2011298}. This emphasis on the importance of social interactions for learning is probably what led Bruner to conceptualize the notion of pragmatic frames \citep{bruner85pragframes}, which has later been reused for example as as conceptual tool to theorize language development \citep{pragmatic_frames_lang_acquisition}. In this work we intent to further motivate the relevance of this notion to enable further progress in Deep RL and AI. 

\paragraph{Human-Robot Interaction} Interactions with knowledgeable human teachers is a well studied form of social interaction. Many works within the Human-Robot Interaction (HRI) and the Interactive learning field studied how to provide interactive teaching signals to their agents, e.g. providing instructions \citep{grizou-2014-hri-instructions}, demonstrations \citep{argall-LfD-2009,grollman-fail-demo-2011}, or corrective advice \citep{celemin-advice}.  In \citep{pragmatic-frames}, authors review this field, showing that many of the considered interaction protocols can be reduced to a restricted set of pragmatic frames. They note that most of these works consider single rigid pragmatic frames. Echoing this observation, the SocialAI benchmark invites to study a broader set of social situations, e.g. requiring agents to both move and speak, and even to learn to interact in a \textit{diversity} of pragmatic frames. Catalysing research on DRL and social skills seems even more relevant now that many application-oriented works are beginning to leverage RL and DRL into real world humanoid social robots \citep{socialRobots2021}.

\paragraph{Recent works on language grounded DRL} Building on NLP, developmental robotics, and previous works on classical goal-conditioned DRL \citep{colas-goal-cond-2020}, a renewed interest emerged towards the development of embodied autonomous agents able to process language \citep{survey-rl-nlp}. Most approaches were proposed to design language conditioned agents in instruction following scenarios (e.g. "go to the red bed"). In \citet{Hermann-grounded-lang-2017}, authors train a DRL model from pixels in a 3D world by augmenting instruction following with auxiliary tasks (language prediction and temporal autoencoding), enabling their agent to follow object-relative instructions ("pick the red object next to the green object"). Multiple other approaches were studied to ease learning: leveraging pre-trained language models \citep{hill2020b-instr-fol-transfer}, demonstrations \citep{fu2018from,Lynch2020ground-lang-in-play} (from which rewards can be learned \citep{BahdanauHLHHKG19}), or using descriptive feedbacks \citep{Nguyen-activity-description-2021} (which has also been tested in combination with imagining new goals \citep{colas-goal-cond-2020}). \citet{Hill2020Environmental} try to assess to which extent vanilla language conditioned agents are able to perform systematic generalization (combining known concepts/skills into new ways). In Embodied Visual Question Answering works, agents are conditionned on questions, requiring them to navigate within environments and then produce an answer ("what color is the bed ?") \citep{eqa,DBLP:conf/cvpr/GordonKRRFF18}. Compared to these previous works, SocialAI aims to enlarge the set of considered scenarios by studying how language conditioned agents are able to ground and produce language within diverse forms of social interactions among embodied social peers.
Closely related to the present work is \cite{imitating-intel},
which is an invitation to focus research on building embodied multi-modal agents suited for human-robot interactions in the real-world. Towards this goal, authors propose a series of experiments on a simulated 3D playroom environment designed for multi-agent interactive scenarios featuring both autonomous agents and/or human players.
The main focus of the paper is in proposing novel ways to leverage human demonstrations to bootstrap agents performance and allow meaningful interactive sessions with humans (as scaffolding a randomly acting agent is a tedious journey).
Because of the complexity of their considered experiments (imitation learning, 3D environments, pixel-based, human in the loop, ...), their work only considers the two now-common social interaction scenarios: Visual Question Answering and instruction following. The novelty of their setup is that these questions/instructions are alternatively produced or tackled by learning agents in an interactive fashion. In SocialAI, we focus on a lighter experimental pipeline (2D grid-world, low dimensional symbolic pixels, no humans) such that we are able to study a broader range of social scenarios, requiring multi-steps conversations and interactions with multiple (scripted) agents within a single episode.

\paragraph{Benchmarks on embodied agents and language}
Multiple benchmarks featuring language and embodied agents already exists.
The BabyAI \citep{chevalierboisvert2018babyai} and gSCAN benchmarks \citep{RuisBench2020} test language conditioned agents on grid-world environments. BabyAI focuses on assessing the sample efficiency of tested agents while gSCAN targets systematic generalization. \cite{MisraBBNSA18} extends this type of benchmark to 3D environments. 
In contrary to SocialAI, these benchmarks do not consider multi-modal action spaces, i.e. agents do not produce language. Besides, they only consider a single rigid social interaction protocol: instruction following.

Related to instruction following benchmarks are testbeds for embodied visual question answering \citep{DBLP:conf/cvpr/GordonKRRFF18,eqa}. Here agents are conditioned on questions: they must navigate within an environment to collect information and produce an answer (i.e. a one or two word output).

\citet{puig2020watchandhelp} propose a new benchmark to test social perception in machine learning models. Learning agents must infer the intent of a scripted agent in a 3D world (using a single demo) to better collaborate towards a similar goal in a new environment. Here again, despite being novel and relevant, only a single social interaction is considered: cooperation towards a common goal, which in that case does not require language use nor understanding.

In between classical disembodied NLP testbeds \citep{CLEVR-2017,wang-etal-2018-glue,social-IQ-2019} and previously discussed embodied language benchmarks is the LIGHT environment \citep{UrbanekLIGHT}, A multiplayer text adventure game allowing to study social settings requiring complex dialogue production \citep{how-to-dragon-2020,fantasy-text-world-2020}. While they consider a text world, i.e. a virtual embodiment, the SocialAI benchmark tackles the arguably harder and richer setting of egocentric embodiment among embodied social peers. Text worlds have also been used in combination with an embodied environment to demonstrate how language-based planning (in text worlds) can benefit instruction following \citep{shridhar2021alfworld}.

Within the Multi Agent Reinforcement Learning field, \cite{DBLP:conf/aaai/MordatchA18} propose embodied environments to study the emergence of grounded compositional language. Here language is merely a discrete set of abstract symbols that can only be used one at a time (per step) and whose meanings must be negociated by agents. While symbol negociation is an intersting social situation to study, we leave it to future work and consider scenarios in which agents must enter an already existing social world (using non-trivial language). In \cite{jacques2019-socialinfl}, authors present Multi Agent social dilemna environments requiring the emergence of cooperative behaviors through communication. In this work communication is stricly non-verbal, while we consider both non-verbal communication (e.g. gaze following) and language based communication.

\section{Environment details}
\label{app:env-details}
\subsection{Action space}
The action space of the environment consists of two modalities (\textit{primitive actions} and \textit{language}) which results in a 3D discrete action vector.

The first dimension corresponds to the primitive action modality, which are identical to actions available in minigrid (\citep{gym_minigrid}, from which our code is based on. It consists of 7 actions (turn left, turn right, move forward, pickup, drop, toggle, done).

In all the environments \textit{pickup} and \textit{drop} actions do not do anything and \textit{done} terminates the episode. We kept these actions as we intend to use them in future iterations of the benchmark.
In TalkItOut, Dance, and CoinThief \textit{toggle} terminates the episode with 0 reward. In DiverseExit, ShowMe and Help \textit{toggle} opens doors and presses buttons. In DiverseExit, it can also be used to \textit{poke} the NPC.

In Dance and CoinThief, only a subset of 3 primitive actions are available (to simplify these environments): \textit{turn left}, \textit{turn right} and \textit{move forward}.

In SocialEnv, all actions behave as in the original environment that is sampled for each new episode.

The second and third dimensions regard the language modality.
The second dimension selects a template and the third a noun. The full grammar for each environments is shown in table \ref{tab:grammar}. In SocialEnv all grammars are merged to a single big one containing all templates and nouns used in all single environments.

Both modalities can also be undefined, i.e. no action is taken in the undefined modality. Examples of such actions are shown in table \ref{tab:action_examples}.

\begin{table}
\centering
\caption{Template-based grammars for all the environments}
\vspace{0.1cm}
\centering
\begin{tabular}{@{}lllll@{}}
\toprule
\textbf{Templates} \\
\toprule 
Action & Template \\
\midrule
 &  TalkItOut, DiverseExit  & CoinThief      & Dance, ShowMe, Help  & SocialEnv \\
\midrule
0 & Where is <noun>         & Here is <noun> & Move your <noun>     & Where is <noun>           \\
1 & Open <noun>             &                & Shake your <noun>    & Open <noun>           \\
2 & Which is <noun>            &                &                      & Close <noun>           \\
3 & How are <noun>          &                &                      & How are <noun>          \\
4 &                         &                &                      & Move your <noun>          \\
5 &                         &                &                      & Shake your <noun>          \\
6 &                         &                &                      & Here is <noun>          \\
7 &                         &                &                      & Which is <noun>          \\
\bottomrule

\end{tabular}
\vspace{1cm}

\begin{tabular}{@{}llllll@{}}
\toprule
\textbf{Nouns} \\
\toprule
Action      & Noun          \\
\midrule
            & TalkItOut     & DiverseExit      &  CoinThief & Dance, ShowMe, Help & SocialEnv \\
\midrule                                         
0           & sesame        & sesame           & 1          & body                & sesame                \\
1           & the exit      & the exit         & 2          & head                & the exit              \\
2           & the wall      & the correct door & 3          &                     & the wall              \\
3           & you           & you              & 4          &                     & the floor             \\
4           & the ceiling   & the ceiling      & 5          &                     & the ceiling           \\
5           & the window    & the window       & 6          &                     & the window            \\
6           & the entrance  & the entrance     &            &                     & the entrance          \\
7           & the closet    & the closet       &            &                     & the closet            \\
8           & the drawer    & the drawer       &            &                     & the drawer            \\
9           & the fridge    & the fridge       &            &                     & the fridge            \\
10          & oven          & oven             &            &                     & oven                  \\
11          & the lamp      & the lamp         &            &                     & the lamp              \\
12          & the trash can & the trash can    &            &                     & the trash can         \\
13          & the chair     & the chair        &            &                     & the chair             \\
14          & the bed       & the bed          &            &                     & the bed               \\
15          & the sofa      & the sofa         &            &                     & the sofa              \\
16          &               &                  &            &                     & the correct door      \\
16          &               &                  &            &                     & 1 \\
17          &               &                  &            &                     & 2 \\
18          &               &                  &            &                     & 3 \\
19          &               &                  &            &                     & 4 \\
20          &               &                  &            &                     & 5 \\
21          &               &                  &            &                     & 6 \\
22          &               &                  &            &                     & body \\
23          &               &                  &            &                     & head \\
\bottomrule

\end{tabular}
\label{tab:grammar}
\end{table}

\begin{table}
\centering
\caption{Examples of various actions in the environment. Second and third dimension must both either be underfined or not.}
\vspace{0.1cm}
\centering
\begin{tabular}{@{}lll@{}}
\toprule
Action & description \\ \midrule
(1, -, -) &  moves left without speaking   \\
(1, 1, 5) & moves left and utters "Open the window" \\
(-, 1, 5) & doesn't move but utters "Open the window" \\
(-, -, -) & nothing happens \\
\bottomrule
\end{tabular}
\label{tab:action_examples}
\end{table}

\subsection{Observation space}

The multimodal observation space consists of the \textit{vision} modality and the \textit{language} modality. 

The \textit{vision} modality is manifested as a \textit{7x7} grid displaying the space in front of the agent (shown as highlighted grids in figure \ref{fig:all_envs}). 
Each location of this grid is encoded as $4$ integers for the object type, color, status and orientation (used for NPCs). \textit{status} is used to refer to object states (e.g. door is \textit{open}) or, if the object is an NPC, it is used to inform about the NPC type (e.g. \textit{wizard} NPC. For example, a blue wizard NPC facing down will be encoded as $(11, 2, 0, 1)$ and a blue guide NPC facing up will be encoded as $(11, 2, 1, 3)$. 

The \textit{language} modality is represented as a string containing the currently heard utterances, i.e. utterances uttered by NPCs next to the agent, and their names (ex. "John: go to the green door"). In case of silence, an "empty indicator" symbol is used.

As it is often more convenient to concatenate all the utterances heard, to simplify the implementation of the agent, the implementation of the environment also supports giving the full history of heard utterances with the "empty indicator" symbols removed as additional information. 

\subsection{The reward}
In all of our environments the extrinsic reward is given upon completing the task. The reward is calculated by the following equation:
\begin{equation}
r_{extr} = 1.0 - 0.9 * \frac{t}{t_{max}}
\label{eq:extr_reward}
\end{equation}
, where $t$ is the number of steps the agent made in the environment and $t_{max}$ is the maximum allowed number of steps.

\subsection{TalkItOut}



This environment consists of three NPCs and four doors, and the goal of the agent is to exit the room using the correct (one out of four) door in $t_{max}=100$ steps.
The agent can find out which door is the correct one by asking the true guide. 
To find out which guide is the correct one, the agent has to ask the wizard.
Before talking to any NPC, the agent has to stand in front of it and introduce himself by saying "How are you?".
Upon finding out which door is the correct one, the agent has to stand in front of it and utter "Open sesame".
Then the episode ends, and the reward is given.

If the agent executes \textit{done}, \textit{toggle} or utters "Open sesame" in front of the wrong door the episode ends with no reward.

An example of a dialogue that might appear in a successful episode is shown in table \ref{tab:successful_episode}
\begin{table}
\centering
\centering
\caption{An example of a successful episode in the TalkItOut environemnt}
\vspace{0.1cm}
\centering
\begin{tabular}{l}
\toprule
True guide: John \\
Correct door color: blue \\
\midrule
\textit{agent goes to the wizard} \\
\textbf{Agent}: How are you? \\
\textbf{Wizard}: I am fine. \\
\textbf{Agent}: Where is the exit? \\
\textbf{Wizard}: Ask John. \\
\textit{agent goes to one guide} \\
\textbf{Agent}: How are you? \\
\textbf{Jack}: I am fine. \\
\textbf{Agent}: Where is the exit? \\
\textbf{Jack}: Go to the red door. \\
\textit{agent goes to the other guide} \\
\textbf{Agent}: How are you? \\
\textbf{John}: I am fine. \\
\textbf{Agent}: Where is the exit? \\
\textbf{John}: Go to the blue door. \\
\textit{agent goes to the blue door} \\
\textbf{Agent}: Open sesame \\
\end{tabular}
\label{tab:successful_episode}
\end{table}

For each episode the colors of doors and NPCs are selected randomly from a set of six and the names of the two guides are selected randomly from a set of two (Jack, John), i.e. in one episode either Jack or John will be the truth speaking guide and the other will be the lying guide.
Furthermore, the grid width and height are randomized from the minimal size of 5 up to 8 and the NPCs and the agent are placed randomly inside (omitting locations in front of doors).


\paragraph{Required social skills}
\label{app:talkitout_social_skills}

In the remainer of this section we use the TalkItOut environment to provide an in-depth example of the detailed list of social skills required in one of our environments (we revert to more straightforward descriptions for all subsequent environment descriptions).

\textit{Intertwined multimodality - }
To solve TalkItOut the agent must use both modalities both in the action and in the observation space.
Furthermore, this multimodality is intertwined because the progression in which the modalities are used is non-trivial.
To discuss this notion further, let's imagine an example of instruction following.
The progression of modalities here is trivial because the agent always \textit{listens} for the command first and then \textit{looks} and \textit{moves/acts} to complete the task.
Another good example is embodied question answering. Here the agent again always first \textit{listens} to the question, then \textit{looks} and \textit{moves} in the environment to finally, at the end, \textit{speak} the answer.

In our environment, however, the agent must always choose which modality to use based on the current state.
Furthermore, it will often be required to switch between modalities many times.
For example, to talk to an NPC the agent first \textit{looks} to find the NPC, then it \textit{moves} to the NPC, finally the agent \textit{speaks} to it and \textit{listens} to the response.
This progression is then used, if needed, for other NPCs, and finally a similar one used to go to the correct door and open it.
Furthermore, depending on the current configuration of environment, the progression can also be different.
Usually, after finding out the correct door the agent needs to \textit{look} for it and \textit{move} to it to \textit{speak} the password, but if the true guide is already next to the correct door only \textit{looking} for the door and \textit{speaking} the password is required.

\textit{Theory of Mind - }
Since the agent must be able to infer good or bad intentions of other NPCs, a basic form of ToM is needed.
Primarily, the agent needs to infer that the wizard is well-intended, wants to help, and is therefore trustworthy.
Using the inferred trust in the wizard, it is possible to infer the good intentions of the true guide, and likewise the bad intentions of the false guide.

On the other hand, as the false guide chooses which false direction to give each time asked, it is also possible to infer its ill-intentions by asking him many times in the same episode and observing this inconsistency.
If an NPC gives different answers for the same question in the same episode, then it is evident its intentions are bad.

\textit{Pragmatic frames - }
Pragmatic frames were not the focus of this environment, and are studied in more detail in other environments, they are present in this environment only in a simple form.
To talk with an NPC the agent needs to stand next to it and introduce itself by saying "how are you", and to get an answer the agent needs to ask "where is the exit".
These simple rules (a.k.a. social conventions) are pragmatic frames, i.e. grammars describing possible and impossible interactions.
For example, it is impossible to communicate if you are far and get directions if you ask "Where is the floor".
The agent needs to be able to extract these rules and use them in relation to all NPCs.

\subsection{Dance}

A Dancer NPC demonstrates a 3-steps dance pattern (randomly generated for each episode) and then asks the agent to reproduce this dance. Each dance step is composed of a primitive action, randomly selected among rotating left, right, or moving forward. $50\%$ of the time, a randomly selected utterance among $4$ possible ones (see table \ref{tab:grammar}) is also performed simultaneously with the primitive action. In the first step of each episode, the NPC utters "Look at me!". It then performs the dance in the next 3 steps. Finally, at the fifth step, the NPC utters "Now repeat my moves", and starts to record the agents' actions. In contrary to TalkItOut, the agent does not need to be close to the NPC to interact with it (i.e. both are "shouting"). To solve \textit{Dance}, the agent must reproduce the full dance sequence. Multiple trials are authorized within the $t_{max}=20$ steps of an episode. Only trials performed after the NPC completed his dance are recorded.

The Dance environment requires the agent to be able to infer that the NPC is setting up a \textit{teaching pragmatic frame} ("Look at me!" + \texttt{do\_dance\_performance} + "Now repeat my moves!"), requiring the agent to \textit{imitate} a social peer, process \textit{multi-modal observations} and produce \textit{multi-modal actions}.
 
\subsection{CoinThief}

 In a room containing 6 coins (randomly placed), a Thief NPC spawns near the agent, i.e. in one of the 4 cells adjacent to the agent (selected randomly for each new episode). At step $0$, the Thief NPC utters "Freeze! Give me all your coins!". The agent can "give coins" by uttering "here is $<nb>$", with $<nb>$ ranging from 0 to 6 (see table \ref{tab:grammar}. Note that the agent does not need to collect coins by navigating within the environment, it only has to utter. To obtain a positive reward, the agent must give exactly the number of coins \textit{that the thief can see}. The thief field of view is a 5x5 square, i.e. a smaller version than the agent's. In addition of its initial orientation facing the agent, the thief also "look around" in another direction, either left or right (selected at random for each episode). Episodes are aborted without reward if the agent use the \textit{move forward} action (the thief wants the agent not to move), or if the maximum number of steps ($t_{max}=20$) is reached. 
 Solving the CoinThief environment requires \textit{Theory of Mind} as the agent must understand that the thief holds \textit{false belief} over the agent's total number of coins and must infer how many coins he actually sees. To infer how many coins the thief sees, the agent must learn the thief's field of view and use memory to remember the thief's two view directions.

\subsection{DiverseExit}
The goal of the agent is to exit the room using one of four doors in $t_{max}=50$ steps. One NPC is present in the environment.
Colors and the initial positions of the NPC and the doors are randomized each episode.

To find out which door is the correct one, the agent has to ask the NPC.
To talk to the NPC, the agent has to introduce himself by saying one of two possible \textit{introductory utterances} (``Where is the exit?'' or ``Which is the correct door?'').

There are 12 different NPC types, one of which is randomly selected to be present in the episode.
Each of the 12 NPCs prefer to be introduced to differently, to be more precise, when the agent utters one of the two \textit{introductory utterances} for the first time in the episode, the \textit{introductory configuration} is saved.
The \textit{introductory configuration} is manifested as the tuple of the following four elements:
(is the agent next to the NPC, was the NPC poked, is eye contact established, which \textit{introductory utterance} was used).
Each NPC must be asked with its preferred \textit{introductory configuration}.
This enables us to create twelve different NPC and their corresponding \textit{introductory configurations}. 
Those twelve configurations are listed in the table \ref{tab:introductory_configurations}. 

If the \textit{introductory configuration} is the one corresponding to the present NPC, the NPC will give the agent the directions (ex. "go to the green door") every time they establish eye contact.
However, if the \textit{introductory configuration} was not the right one, the NPC will not give the directions in this episode (a once saved introductory state cannot be overwritten in the same episode). 

To solve TeachingGames the agent must learn a large diversity of different frames (12). Furthermore, it must learn to differentiate between them and infer which frame to use with which NPC.

\begin{table}
\centering
\caption{Twelve \textit{introductory configurations} corresponding to twelve possible different NPCs in DiverseExit.}
\vspace{0.1cm}
\centering
\begin{tabular}{@{}lllll@{}}
\toprule
npc\_type   & is next to the NPC & was the NPC poked    & eye contact & introductory utterance used \\ \midrule
0           & next to            & poked                & Yes                          & ``Where is the exit'' \\
1           & next to            & not poked            & Yes                          & ``Where is the exit'' \\
2           & not next to        & not poked            & Yes                          & ``Where is the exit'' \\
3           & next to            & poked                & Yes                          & ``Which is the correct door`` \\
4           & next to            & not poked            & Yes                          & ``Which is the correct door`` \\
5           & not next to        & not poked            & Yes                          & ``Which is the correct door`` \\
6           & next to            & poked                & No                           & ``Where is the exit'' \\
7           & next to            & not poked            & No                           & ``Where is the exit'' \\
8           & not next to        & not poked            & No                           & ``Where is the exit'' \\
9           & next to            & poked                & No                           & ``Which is the correct door`` \\
10          & next to            & not poked            & No                           & ``Which is the correct door`` \\
11          & not next to        & not poked            & No                           & ``Which is the correct door`` \\
\bottomrule
\end{tabular}
\label{tab:introductory_configurations}
\end{table}

\subsection{ShowMe}
The goal of the agent is to exit the room through the door placed at the top wall of the environment in $t_{max}=100$ steps.
At the beginning of the episode the door is locked and can be unlocked by activating the correct switch (one out of three) on the bottom wall.
The switches can be activated using the \textit{toggle} action, however once a switch is activated it cannot be deactivated.
This means that the agent must press the correct switch from the first try.
The information about which switch is the correct one can be inferred by looking at the NPC.
Once eye contact is established with the NPC, the NPC will say "Look at me" and proceed to press the switch and exit the room.
After this, the switches are reset and the door is locked once again.
As the switch doesn't change whether it's activated or not (it looks the same) the agent must infer what switch was pressed by looking at the NPCs movement and infer which switch was activated.
The reward is given once both the NPC and the agent have left the room, i.e. if the agent leaves the room before the NPC no reward is given.

To solve ShowMe, the agent must learn to imitate the NPCs goals (pressing the correct switch) from its behavior (a facet of ToM).
Furthermore, it has to infer the \textit{Teaching pragmatic frame} where the \textit{slot} is the pressed button.

\subsection{Help}
The environment consists of two roles: The Exiter and the Helper. The shared goal of both the participants it for the Exiter to exit the environment in $t_{max}=20$ steps. It can do so using any of the two doors on the right wall of the environment. At the beginning of the episode both doors are locked and can be unlocked by pressing the corresponding switch on the left wall. The environment is separated by a wall of lava in the middle, disabling movement from the left to the right side of the room. The Exiter is placed on the right side of the environment (next to the doors) and the Helper on the left side (next to the switches). As the episode ends without reward if both switches are pressed, the two participants have to agree on which door to use.

The purpose of this environment is to train the agent in the Exiter role and test in the Helper role.

When the agent is in the Exiter role (training phase), the NPC is in the Helper role. Then the NPC acts as follows. It moves towards the switch corresponding to the door that is the closest to the agent. Once in front of the switch, it looks at the agent and waits for eye contact. Once eye contact has been established, the NPC activates the switch. The agent, therefore, needs to learn to choose a door and confirm this choice by establishing eye contact.

When the agent is in the Helper role (testing phase), the NPC is in the role of the Exiter. Then the NPC chooses a door and moves in front of it. Once there, it looks at the agent and waits for eye contact. Once eye contact has been established, the NPC attempts to exit the room by the door.

To solve Help, the agent must learn the whole pragmatic frame just from seeing its own perspective of it. It must learn to infer the shared goal and which actions from both the agent and the NPC lead to the achievement of this goal.

\subsection{SocialEnv}

SocialEnv is a \textit{meta-environment}, i.e. it is a multi-task environment in which, for each new episode, the agent is randomly spawned into one of the 6 previously discussed environments. The agent's grammar is a set of all previous grammars (see table \ref{tab:grammar}). $t_{max}$ is set to the original $t_{max}$ of each environment.

Solving SocialEnv requires to infer what is the current social scenario (i.e. the current environnment) the agent is spawned in. This can be achieved by leveraging pragmatic information collected through interaction, i.e. differentiating environments from social interaction footprints. For instance, a proficient agent could reliably detect it is in the TalkItOut environment by observing that there are 3 NPCs (1 wizard-type and 2 guide-type). Given that this environment detection is mastered, the agent still has to be proficient in \textit{all of the core social skills} we proposed, to be able to solve each environment.

\section{Experimental details}
\label{app:experimental_details}

\subsection{Baselines details}
\label{app:cond-details}

\begin{figure}
    \centering
    \includegraphics[width=0.4\columnwidth]{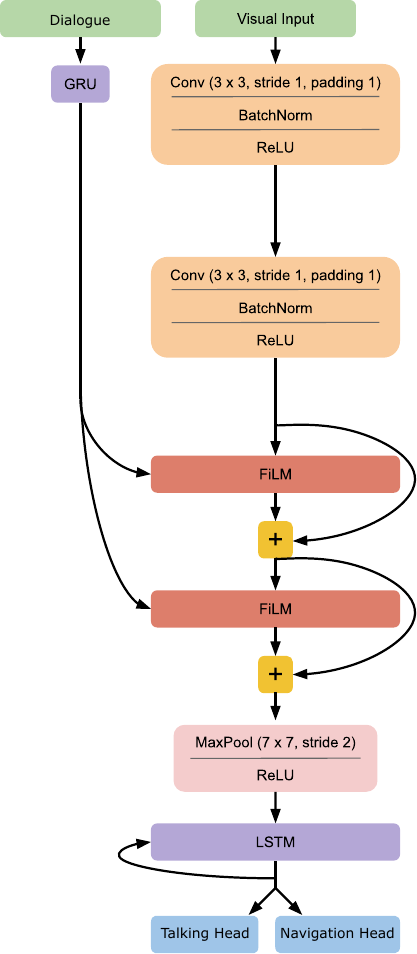}
    \caption{Our Multi-Headed PPO baseline DRL agent. Architecture visualization is a modified version of the one made by \citet{hui2020babyai}. We perform two modifications: 1) Instead of fixed instruction inputs our model is fed with NPC's language outputs (if the agent is near an NPC), and 2) We add a language action head, as our agent can both navigate and talk.}
    \label{fig:baby-model}
\end{figure}

\paragraph{PPO baseline} In this work we use a PPO-trained \citep{ppo} DRL architecture initially designed for the BabyAI benchmark \citep{chevalierboisvert2018babyai}. The policy design was improved in a follow-up paper by \citet{hui2020babyai} (more precisely, we use their \textit{original\_endpool\_res} model). See figure \ref{fig:baby-model} for a visualization of the complete architecture. First, symbolic pixel grid observations are fed into two convolutional layers \citep{lecun1989backpropagation, krizhevsky2012imagenet} (3x3 filter, stride and padding set to 1), while dialogue inputs are processed using a Gated Recurrent Unit layer \citep{gru}. The resulting image and language embeddings are combined using two FiLM attention layers \citep{film}. Max pooling is performed on the resulting combined embedding before being fed into an LSTM \citep{lstm} with a $128D$ memory vector.
The LSTM embedding is then used as input for the navigation action head, which is a two-layered fully-connected network with tanh activations and has an 8D output (i.e. 7 navigation actions and no\_op action).

In order for our agent to be able to both move and talk, we add to this architecture a talking action head, which is composed of three networks. All of them are two-layered, fully-connected networks with tanh activations, and take the LSTM's embedding as input. The first one is used as a switch: it has a one-dimensional output to choose whether the agent talks (output > 0.5) or not (output < 0.5). If the agent talks, the two other networks are used to respectively sample the template and the word.

Note that the textual input given to the agent consists of the full dialogue history (without the "empty string" indicator) as we found it works better than giving only current utterances.

\begin{table}
\centering
\caption{Training hyperparameters}
\vspace{0.1cm}
\centering
\begin{tabular}{@{}lll@{}}
\toprule
Hyperparameter & value \\ \midrule
learning rate       & $1 e^4$    \\
GAE $\lambda$       & $0.99$    \\
clip $\epsilon$     & $1 e^5$    \\
batch size          & $1280$    \\
$\gamma$            & $0.99$   \\
recurrence          & $10$ \\
epochs              & $4$  \\
\bottomrule
\end{tabular}
\label{tab:train_hyperparmas}
\end{table}

\begin{table}
\centering
\caption{Exploration bonus hyperparameters}
\vspace{0.1cm}
\centering
\begin{tabular}{@{}llllllll@{}}
\toprule
Hyperparameter & value \\ \midrule
            & TalkItOut & Dance & CoinThief & DiverseExit   & ShowMe    & Help      & SocialEnv  \\ \midrule
type        & lang  &  vision   & vision  & lang          & vision    & vision    & vision     \\
T           & $0.6$ &  $0.6$    & $0.6$ & $0.6$         & $0.6$     & $0.6$     &  $0.6$      \\
C           & $7$   &   $3$        & $2$ & $20$          & $3$       & $3$       &    $2$        \\
M           & $50$  &   $50$        & $50$ & $50$          & $50$      & $50$      &    $50$        \\
\bottomrule
\end{tabular}
\label{tab:explo_hyperparmas}
\end{table}

\paragraph{Exploration bonus} 
The exploration bonus we use is inspired by recent works in intrinsically motivated exploration \citep{pathakICMl17curiosity, curiositythroughreachability, tang2017exploration}.
These intrinsic rewards estimate the novelty of the currently observed state and add the novelty based bonus to the extrinsic reward. The novelty is estimated by counting various aspects of the state. We make our reward episodic by resetting the counts at the end of each episode.

In this work we study two different techniques for computing the exploration bonus (counting), and we use the one that was more suitable for a given environment. Which reward was used for which environment and the corresponding hyperparameters are visible in table \ref{tab:explo_hyperparmas}. The two different techniques are: language-based and vision-based.

In the language-based intrinsic reward we count how many times was each utterance observed and compute an additional bonus based on the following equation:
\begin{equation}
r_{intr} = tanh(\frac{C}{(N(s_{lang})+1)^{M})*T)})
\label{eq:expl_bonus_lang}
\end{equation}
, where $M$, $C$, and $T$ are hyperparameters and $N(s_{lang})$ is the number of times the utterance $s_{lang}$ was observed during this episode.
In the current version of the environment the agent cannot hear his own utterances and the NPCs speak only when spoken to.
Therefore, this exploration bonus can be seen as analogous to social influence \citep{jacques2019-socialinfl} in the language modality, as the reward is given upon \textit{making the NPC respond}. 

In the vision-based intrinsic reward, we reward the agent for seeing a new encoding. An encoding is the 4D representation of a grid (object\_type, color, additional\_information, orientation). At each step, a set of encountered encodings is created by removing the duplicates, and then the reward computed by the following equation:
\begin{equation}
r_{intr} = tanh(\sum_{e \in E(s)} \frac{C}{(N(e)+1)^{M})*T} )
\label{eq:expl_bonus_cell}
\end{equation}
, where $M$, $C$, and $T$ are as in equation \ref{eq:expl_bonus_lang}, $E(s)$ is a set of unique encodings visible in state $s$, and $N(e)$ is the number of times an encoding $e$ was encountered.

These intrinsic rewards are a good example of biases that have to be discovered for training social agents.

\subsection{Computational ressources}

To perform our experiments, we used a slurm-based cluster. Producing our final results require to run 16 seeds of 3 different conditions on each of our 10 environments (7 environments and 3 modified environments for our case-studies), i.e. 480 seeds. Each of these experiments takes approximately 42 hours on one CPU and one 32GB Tesla V100 GPU (one GPU can serve 4 experiments in parallel), which amounts to $20160$ CPU hours and $5040$ GPU hours.

\end{document}